\documentclass{article}

\usepackage{microtype}
\usepackage{graphicx}
\usepackage{subfigure}
\usepackage{booktabs} %

\usepackage{hyperref}

\usepackage{booktabs}
\usepackage{colortbl}

\usepackage{graphicx}
\usepackage{float,wrapfig}

\usepackage{array}
\usepackage{adjustbox}
\usepackage{multirow}
\usepackage{colortbl}

\usepackage{caption}
\usepackage{color,xcolor}
\usepackage{epsfig}
\usepackage{bbm}
\usepackage{hhline}

\usepackage{amsmath,amsfonts,amsthm,amssymb}
\usepackage{mathtools}
\usepackage{bm}
\usepackage{nicefrac}
\usepackage{microtype}

\usepackage{changepage}
\usepackage{extramarks}
\usepackage{fancyhdr}
\usepackage{lastpage}
\usepackage{setspace}
\usepackage{soul}
\usepackage{xspace}

\usepackage{url}
\usepackage{algorithm, algorithmic}
\usepackage{todonotes} %

\usepackage{titlesec}

\DeclareMathOperator*{\argmin}{arg\,min}

\newcommand{\sect}[1]{Section~\ref{#1}}

\newcommand{\eqn}[1]{Equation~\ref{#1}}
\newcommand{\fig}[1]{Figure~\ref{#1}}
\newcommand{\tbl}[1]{Table~\ref{#1}}
\newcommand{\myparagraph}[1]{{\bf #1}\quad}

\DeclarePairedDelimiterX{\infdivx}[2]{(}{)}{%
  #1\;\delimsize|\delimsize|\;#2%
}

\newcommand{\model}{IRED\xspace}

\makeatletter
\DeclareRobustCommand\onedot{\futurelet\@let@token\@onedot}
\def\@onedot{\ifx\@let@token.\else.\null\fi\xspace}

\def\eg{\emph{e.g}\onedot} 
\def\ie{\emph{i.e}\onedot}

\makeatother

\definecolor{MyDarkBlue}{rgb}{0,0.08,1}
\definecolor{MyDarkGreen}{rgb}{0.02,0.6,0.02}
\definecolor{MyDarkRed}{rgb}{0.8,0.02,0.02}
\definecolor{MyDarkOrange}{rgb}{0.40,0.2,0.02}
\definecolor{MyPurple}{RGB}{111,0,255}
\definecolor{MyRed}{rgb}{1.0,0.0,0.0}
\definecolor{MyGold}{rgb}{0.75,0.6,0.12}
\definecolor{MyDarkgray}{rgb}{0.66, 0.66, 0.66}

\newcommand{\myitem}{\vspace{-5pt}\item}

\usepackage{amsmath,amsfonts,bm}

\def\eqref#1{equation~\ref{#1}}

\def\1{\bm{1}}

\def\vx{{\bm{x}}}
\def\vy{{\bm{y}}}

\DeclareMathAlphabet{\mathsfit}{\encodingdefault}{\sfdefault}{m}{sl}
\SetMathAlphabet{\mathsfit}{bold}{\encodingdefault}{\sfdefault}{bx}{n}

\usepackage[accepted]{icml2024}

\icmltitlerunning{Iterative Reasoning through Energy Diffusion}

\begin{document}
\twocolumn[
\icmltitle{Learning Iterative Reasoning through Energy Diffusion}

\icmlsetsymbol{equal}{*}

\begin{icmlauthorlist}
\icmlauthor{Yilun Du}{mit,equal}
\icmlauthor{Jiayuan Mao}{mit,equal}
\icmlauthor{Joshua Tenenbaum}{mit}
\end{icmlauthorlist}

\icmlaffiliation{mit}{MIT}

\icmlcorrespondingauthor{Yilun Du}{yilundu@mit.edu}
\icmlcorrespondingauthor{Jiayuan Mao}{jiayuanm@mit.edu}

\icmlkeywords{Diffusion Models, Iterative Reasoning,  Energy Based Models}

\vskip 0.3in
]

\printAffiliationsAndNotice{\icmlEqualContribution} 

\begin{abstract}
We introduce {\it iterative reasoning through energy diffusion} (\model), a novel framework for learning to reason for a variety of tasks by formulating reasoning and decision-making problems with energy-based optimization. \model learns energy functions to represent the constraints between input conditions and desired outputs. After training, \model adapts the number of optimization steps during inference based on problem difficulty, enabling it to solve problems outside its training distribution --- such as more complex Sudoku puzzles, matrix completion with large value magnitudes, and path finding in larger graphs. Key to our method's success is two novel techniques: learning a sequence of annealed energy landscapes for easier inference and a combination of score function and energy landscape supervision for faster and more stable training. Our experiments show that \model outperforms existing methods in continuous-space reasoning, discrete-space reasoning, and planning tasks, particularly in more challenging scenarios. Code and visualizations are at \url{https://energy-based-model.github.io/ired}.
\end{abstract}

\section{Introduction}

Being able to solve complex reasoning tasks such as logic inference, mathematical proofs, and decision-making is one of the hallmarks of artificial intelligence. Researchers in various fields have been working on domain-specific algorithms for solving these tasks, typically utilizing various forms of search or optimization in iterative manners (\eg, dynamic programming and gradient descent). These domain-specific algorithms are usually highly efficient and effective, but they usually can not directly handle sensory data and typically require users or experts to encode rules in domain-specific languages (such as axioms used in mathematical provers or domain theories in planning). Furthermore, it is usually hard for these systems to learn from experience to improve their performance on familiar tasks. A large body of work has been trying to address these limitations by incorporating machine learning in order to handle sensory inputs and learn to formulate and solve problems. Typical ideas include utilizing these domain-specific solvers as a submodule in a deep neural network~\citep[e.g., SAT solvers;][]{wang2019satnet} or building structured neural networks that can realize algorithms~\citep[e.g., dynamic programming;][]{xu2019can}.

\begin{figure}[t]
\centering
\includegraphics[width=1.0\linewidth]{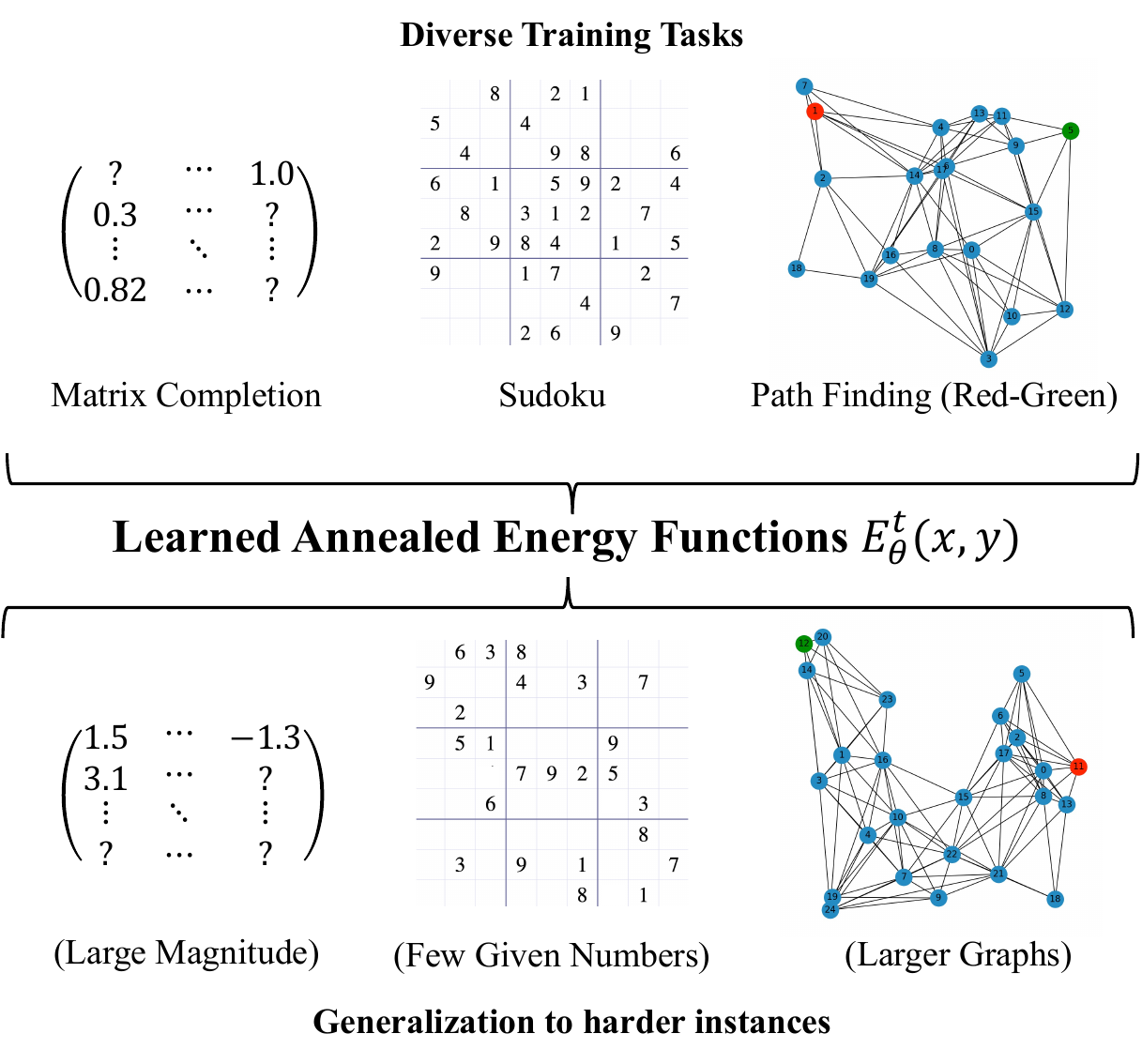}
\vspace{-25pt}
\caption{\small \textbf{Reasoning as Energy Diffusion} -- \model formulates reasoning problem with inputs $\vx$ and output $\vy$, as an energy minimization problem over a learned energy function. It can be trained stably for a wide variety of reasoning tasks and achieves strong generalization to harder problem instances, through adaptive computation in the optimization process.}
\label{fig:teaser}
\vspace{-20pt}
\end{figure}

Illustrated in \fig{fig:teaser}, we take a different approach to address the aforementioned challenges by formulating various kinds of reasoning and decision-making problems as an optimization problem. In particular, we consider the learning-to-reason problem as learning an energy function $E_\theta(\vx, \vy)$ over input conditions $\vx$ and desired output $\vy$. For example, logical deduction can be cast as finding possible assignments to variables that satisfy all logical constraints; theorem proving can be cast as finding a sequence of valid deduction steps that entails the goal; planning can be cast as finding a sequence of actions that respect the transition model of the environment and achieve the goal. This formulation directly allows us to learn the underlying constraints for a given task automatically from input-output data, without additional task-specific knowledge. Therefore, we can solve a wide variety of tasks across different domains using the same underlying training and inference paradigm, by only swapping out the neural network encoder for different data formats of $\vx$ and $\vy$.
Another important feature of this optimization-based formulation is that during inference time, we can choose to apply a different amount of computation depending on the hardness of the problem by inspecting the value of the function $E_\theta(\vx, \vy)$.

In particular, in this paper, we propose {\it iterative reasoning through energy diffusion} (\model), a general framework for learning to reason. \model is trained on a dataset of paired $(\vx, \vy)$ data, and can recover the underlying energy function describing the objective function and constraints. During inference, because we are explicitly solving an optimization problem of finding the $\vy^*$ that maximizes the energy function $E_\theta$, we can run an adaptive number of optimization steps depending on the hardness of the problem. This enables us to solve problems that are beyond the training distribution, for example, Sudoku puzzles with a harder difficulty level, matrix manipulation under worse condition numbers, and sorting arrays with a larger size.

\looseness=-1
Our paper is not the first one to propose the use of energy-based models (EBMs) as a general framework for learning and reasoning~\citep[see, for example,][]{du2022learning}. Although being a general framework for learning and reasoning, existing work falls short in its training speed, stability, and inference-time optimization hardness. These issues are critical and fundamentally hard because the learning of EBMs typically involves back-propagation through the entire iterative optimization process, and in general, the function landscape of $E_\theta$ can be complex with a large number of local optima. In this paper, we propose two important techniques to address these two challenges. Drawing inspiration from diffusion models and their relations to energy-based models~\citep{ho2020denoising,du2023reduce}, instead of learning a single energy landscape, we instead learn a sequence of annealed energy landscapes, where smoother landscapes are being first optimized before optimizing for sharper ones afterward. Furthermore, in contrast to earlier work on EBM learning, \model uses a combination of denoising supervision and direct supervision through negative sample mining. Both techniques can be implemented without the need to backpropagate through the optimization process, thereby making our learning algorithm both stable and fast.

We show the effectiveness of \model on three groups of tasks: continuous-space reasoning (\eg, matrix completion, inversion), discrete-space reasoning (\eg, Sodoku solving, graph connectivity prediction), and planning (\eg, finding paths on graphs). Compared with various domain-specific and domain-independent learning-to-reason baselines, including recurrent adaptive computation~\citep{palm2018recurrent}, EBM~\citep{du2022learning} and diffusion-based models~\citep{ho2020denoising}, \model outperforms all of them, especially on test instances that are of higher difficulty levels, such as on matrices with larger value magnitudes, sudoku of higher difficulty levels, and larger graphs. Ablation studies show that the proposed optimization paradigm enables stable training and better generalization.

\section{Related Work}

\begin{figure*}
    \centering\small
    \includegraphics[width=\textwidth]{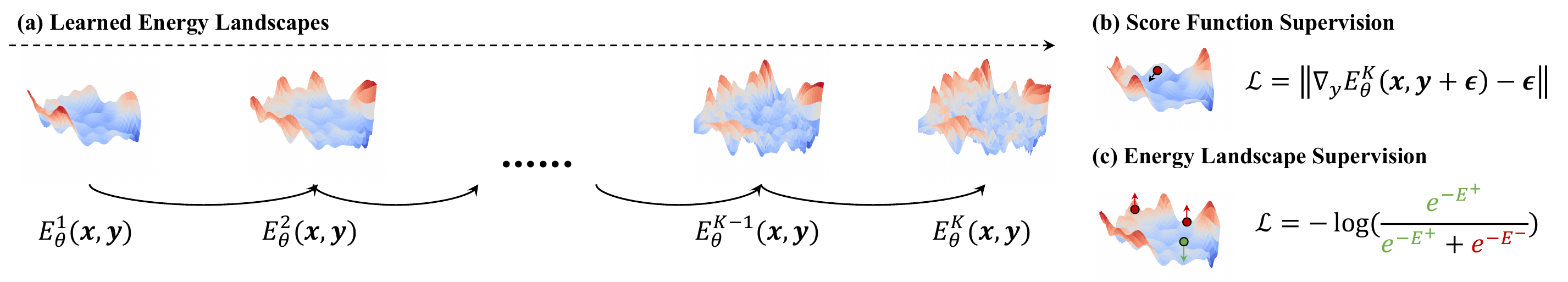}
    \vspace{-20pt}
    \caption{\textbf{\model Learns a Sequence of Energy Landscapes.} During inference time, we optimize for $\vy^*$ that minimizes the energy function, and we gradually increase the complexity of the energy optimization problem. The energy functions are trained with a combination of score function supervision and energy landscape supervision.}
    \label{fig:model}
    \vspace{-15pt}
\end{figure*}

\myparagraph{Learning to reason with optimization.} A wide variety of reasoning problems can be formulated as an optimization problem, including constraint satisfaction problems (CSPs), mathematical programs,  discrete-space~\citep{kautz2006satplan} and continuous-space~\citep[\ie, trajectory optimization, see][for a survey]{bryson2018applied} optimization problems, and even algorithmic reasoning tasks~\citep{brockett1991dynamical}. The high-level idea is to cast these inference and decision-making problems as finding a set of variables that minimizes an objective function subject to constraints. Recently, there has been a growing interest in learning the objective and constraint functions instead of manually specifying them, which would be useful for domains where people do not have expert knowledge or simply the functions are too hard to be specified (\eg, over high-dimensional sensory inputs).

Along this line, the first group of papers has explored using domain-specific optimization solvers as a computation block in neural networks. For example, \citet{amos2017optnet, donti2017task} integrates quadratic program solvers, \citet{djolonga2017differentiable, wilder2019melding} studies submodular programs solvers, \citet{wang2019satnet} uses differentiable Max-SAT solvers, \citet{yang2023neurasp} considers answer-set programming (ASP) solvers, \citet{manhaeve2018deepproblog} considers probabilistic logic programming solvers, and \citet{rocktaschel2017end} integrates symbolic theorem-proving solvers. However, due to the dependence on a particular problem formulation language, these frameworks are usually limited to solving problems of a particular kind.

The second group of papers has explored using a generic optimization framework as the underlying formulation. For example, \citet{bai2019deep, anil2022path} utilizes equilibrium energy minimization inside a neural network to save memory, \citet{rubanova2021constraint, comas2023inferring} utilizes energy minimization to simulate physical dynamics by using neural networks to parameterize an energy function. Our paper falls into this group as well. Similar to our work, \citet{du2022learning} uses energy-based models for learning to reason. In contrast to it, this paper proposes to use a combination of denoising diffusion and supervised energy landscape training. This gives us stable training and strong performance. A concurrent work from \citet{sun2023difusco} also considers using diffusion models for solving combinatorial optimization problems. However, their formulation is developed specifically for combinatorial optimization problems on graphs, but our diffusion formulation with energy parameterizations, landscape supervision, and substep optimizations can be generally applied to many optimization and decision-making domains.

\myparagraph{Learning to reason with iterative neural computation.} Another popular line of research studies using neural networks with iterative computation for reasoning. They draw motivation from the fact that many domain-specific constraint solvers are indeed iterative optimization algorithms (\eg, gradient descent). At a high level, there are two groups of work: leveraging explicit program representations~\citep[][; typically with external memories]{graves2014neural, neelakantan2015neural, reed2015neural, cai2017making, chen2020compositional, banino2021pondernet} and using recurrent neural networks~\citep{graves2016adaptive, kaiser2015neural, chung2016hierarchical, bolukbasi2017adaptive, yang2017differentiable, dong2019neural, dehghani2018universal, schwarzschild2021can,yang2023learning}. One of the key challenges in both types of approaches is when to halt the computation. Researchers have been tackling this problem through reinforcement learning~ \citep{chen2020learning, chung2016hierarchical}, leveraging hierarchical decomposition of programs~\citep{cai2017making}, heuristic policies~\citep{bolukbasi2017adaptive}, and variational inference~\citep{banino2021pondernet}. However, these approaches are usually unstable, and many of them require manual hyper-parameter tuning~\citep{banino2021pondernet} or additional human annotations~\citep{cai2017making}. In this paper, we focus on an orthogonal approach by solving a broad set of reasoning problems by casting them as an optimization on learned energy landscapes. During optimization, the energy function of the landscapes naturally acts as a termination criterion.

\myparagraph{Energy-based models and diffusion models.} Our work is related to past work formulating prediction using Energy-Based Models (EBMs) \citep{lecun2006tutorial}. Most recent EBMs have focused on learning probabilistic models over data \citep{xie2016theory,xie2018cooperative,du2019implicit, grathwohl2020cutting, du2021improved, arbel2020ebm, xiao2020vaebm} but most similar to our work ~\citep{du2022learning} focuses on using energy minimization to solve reasoning tasks.  Our work leverages the connection of energy based models and diffusion models ~\citep{du2023reduce} to more effectively learn energy landscapes for solving reasoning problems. 

An important difference between our proposed approach and standard diffusion models is that diffusion models usually focus on learning a particular sampling path transitioning from Gaussian noise to a target solution, where individual transition kernels across timesteps are learned. However, when obtaining a solution using these transition kernels, errors often accumulate across sampling timesteps, preventing a diffusion model from obtaining a precise answer to a reasoning problem. By contrast, we formulate predicting solutions as optimizing an annealed sequence of energy landscapes. In this setting, multiple steps of optimization are run at each energy landscape to ensure that we are at an energy minima at every landscape. These multiple steps of optimization prevent the accumulation of errors from using transition kernels in diffusion models, as they project the sample to an energy minima, which is likely ``in distribution'' to what has been seen during training.

\section{Learning Iterative Reasoning through Energy Optimization}

Let $\mathcal{D} = \{X, Y\}$ be a dataset for a reasoning task consisting of inputs $\vx \in \mathbb{R}^O$ and corresponding solutions $\vy \in \mathbb{R}^M$. We aim to learn a neural network-based prediction model $\text{NN}_\theta(\cdot)$ which can generalize execution $\text{NN}_\theta(\vx')$ to a test distribution $\vx' \in \mathbb{R}^{O'}$, where $\vx'$ can be significantly larger and more challenging than the training data $\vx \in X$ (\eg, of higher dimensions, or with larger number magnitudes), by leveraging a possibly increased computational budget.

We formulate this adaptive model as an iterative energy optimization in \sect{sect:energy_min}. Our overall framework is illustrated in \fig{fig:model}. In particular, we construct an annealed sequence of energy functions to improve optimization. To involve training stability, speed, and performance, we propose to shape the energy landscape to correctly assign minimal energy to ground truth solutions. We provide full pseudocode for training our approach in \sect{sect:psuedocode} with training following Algorithm \ref{alg:train} and inference following Algorithm \ref{alg:test}.

\begin{figure*}
\begin{minipage}[t]{0.49\textwidth}
\begin{algorithm}[H]
\begin{algorithmic}
    \STATE \textbf{Input:} Problem Dist $p_D(\vx, \vy)$, EBM $E_\theta(\cdot)$, Noise Schedules $\{\sigma_k\}$, Corruption Function $c(\cdot)$, Landscapes $k$.
    \vspace{0.2em}
    \WHILE{not converged}
    \STATE \emph{$\triangleright$ Supervise the Energy Landscape through Denoising:}
    \STATE $\vx_i, \vy_i \sim p_D, \epsilon \sim \mathcal{N}(0, 1), k \sim \{1, \ldots, K\}$
    \STATE $\tilde{\vy_i} \gets \sqrt{1-\sigma_k^2}\vy_i + \sigma_k \epsilon $
    \STATE $\mathcal{L}_{\text{MSE}} \gets \|\nabla_y E_\theta(\vx_i, \tilde{\vy_i}, k)  - \epsilon \|^2$
    \vspace{2mm}
    
    \STATE \emph{$\triangleright$ Shape the Energy Landscape Contrastively:}
    \STATE $\vy^{-}_i \gets c(\vy_i)$
    \STATE $\tilde{\vy_i}^{-} \gets \sqrt{1-\sigma_k^2}\vy^{-}_i + \sigma_k \epsilon $
    \STATE $E^{+}_{i} \leftarrow E_\theta(\vx_i, \tilde{\vy_i}, k)$; $E^{-}_{i} \leftarrow E_\theta(\vx_i, \tilde{\vy}^-_i, k)$
     \STATE $\mathcal{L}_{\text{Contrast}} \leftarrow -\log \left ( \frac{e^{-E^+_i}}{e^{-E^+_i} + e^{-E^-_i}} \right) $

    \vspace{2mm}
    
     \STATE \emph{$\triangleright$ Optimize objective $\mathcal{L}_{\text{MSE}} + \mathcal{L}_{\text{Contrast}}$ wrt $\theta$:} 
    \STATE $\Delta \theta \gets \nabla_\theta (\mathcal{L}_{\text{MSE}} + \mathcal{L}_{\text{Contrast}}) $
    \STATE Update $\theta$ based on $\Delta \theta$ using Adam optimizer 
    
    \ENDWHILE

  \end{algorithmic}
 \caption{\model Training}
 \label{alg:train}
 \end{algorithm}
\end{minipage}
\begin{minipage}[t]{0.49\textwidth}
\begin{algorithm}[H]
\begin{algorithmic}
    \STATE \textbf{Input:} Input task $\vx_i$, Step Sizes $\lambda_k$, Number of Landscapes $K$, EBM $E_\theta(\cdot)$, Optimization Steps $T$.
    \vspace{0.2em}
    \STATE $\tilde{\vy}_i \sim \mathcal{N}(0, 1)$
    \FOR{each landscape $k = 1$ to $K$}
    \FOR{run T steps of optimization $t = 1$ to $T$}
    \STATE \emph{$\triangleright$ Optimize candidate solution $\tilde{\vy}_i$ with gradient:}
    \STATE $\tilde{\vy}_i' \gets \tilde{\vy}_i -  \lambda_k \nabla_\vy E_\theta (\vx_i, \tilde{\vy}_i, k)$
    \vspace{2mm}
    \STATE \emph{$\triangleright$ Check if the gradient descent step decreases energy:}
    \IF{$E_\theta (\vx_i, \tilde{\vy}_i, k) > E_\theta (\vx_i, \tilde{\vy}_i', k)$}
        \STATE $\vy_i \gets \vy_i'$
    \ENDIF
    \ENDFOR
    \vspace{0.8mm}
    \STATE \emph{$\triangleright$ Scale optimized candidate solution:}
    \STATE $\tilde{\vy}_i \gets \frac{\sqrt{1-\sigma_k^2}}{\sqrt{1-\sigma_{k-1}^2}} \tilde{\vy}_i$
    \ENDFOR
    \vspace{2mm}
    \STATE \textbf{return} $\vy = \tilde{\vy}_i$
  \end{algorithmic}
 \caption{\model prediction algorithm}
 \label{alg:test}
 \end{algorithm}
 \end{minipage}
 \vspace{-10pt}
 \end{figure*}

\subsection{Reasoning as Annealed Energy Minimization}
\label{sect:energy_min}

A wide variety of reasoning and decision making problems can be formulated as an optimization problem.  Traditionally, researchers have been focused on designing various domain-specific algorithms for solving different problems, typically with search, gradient-based optimization, or other forms of iterative computation, and also integrating machine learning to help. In this work, we take a different approach of formulating various kinds of reasoning and decision-making problems as an optimization process over a learned energy-based model (EBM): $E_\theta(\vx, \vy): \mathbb{R}^O \times \mathbb{R}^M \rightarrow \mathbb{R}$. Under this formulation, the final prediction problem can be cast as finding the solution $\vy$ according to: \begin{equation}
    \vy = \argmin_{\vy} E_\theta(\vx, \vy).
\end{equation}
One can use gradient descent to find such solutions:
\begin{equation}
   \vy^t = \vy^{t-1} - \lambda \nabla_{\vy} E_\theta(\vx, \vy^{t-1}),
   \label{eqn:energy_iterative}
\end{equation}
where $\lambda$ is the step size for optimization and the initial prediction $\vy^0$ is initialized from a fixed noise distribution (\ie, Gaussian throughout the paper). The final output of $\vy^T$ is obtained after $T$ steps of optimization. 

In earlier work using a similar formulation \citet{du2022learning}, such EBM $E_\theta(\vx, \vy)$ is trained by differentiating through the $T$ steps of optimization and minimizing the MSE with the ground truth label $\vy$
$\mathcal{L}_{\text{Opt}}(\theta) = \| \vy_i^T - \vy_i \|^2.$
This approach requires the forward and backward computation of K steps of optimization at training, which makes it slow and unstable. Furthermore, because the EBM $E_\theta$ may have a complex optimization landscape\footnote{For example, the 3-SAT problem exhibits steep energy minima surrounded by flat energy landscapes.}, robustly finding solutions to \eqn{eqn:energy_iterative} is fundamentally hard.

As a general solution to stable training and better test-time optimization, instead of directly learning $E_\theta(\vx, \vy)$, at a high-level, we propose to learn a sequence of annealed energy functions $E^k_\theta$ ($k=0,1,\cdots,K$), and supervise the EBM learning with the gradient of the energy function:
\begin{equation}
    \label{eqn:denoise}
      \mathcal{L}_{\text{MSE}}(\theta) = \| \nabla_{\vy} E_\theta(\vx, \vy + \epsilon) - \epsilon \|^2, \quad \epsilon \sim \mathcal{N}(0, 1).
\end{equation}
During training time, we obtain the ground truth for $\vy$ by generating a noise-corrupted label $\vy + \epsilon$, following a schedule of noise corruptions. By supervising on the gradient, our approach is substantially faster and more stable than earlier works using plain EBMs~\citep{du2022learning} as it only supervises training of a single step of the optimization.

\subsection{Learning Sequence of Annealed Energy Landscapes}
\label{sect:energy_annealed}

Our key idea to mitigate the hard optimization problem of \eqn{eqn:energy_iterative} is to use simulated annealing --- where smoother energy landscapes are first optimized before optimizing sharper ones afterward, as illustrated in \fig{fig:model}.

Similar to diffusion models~\citep{sohl2015deep, ho2020denoising}, we propose to optimize and learn an annealed sequence of energy landscapes, with earlier energy landscapes being smoother to optimize and the latter ones more difficult. Given a ground truth label $\vy$, we learn a sequence of $K$ energy functions\footnote{Empirically, in our experiments, we found that setting $K=10$ was sufficient across all the domains we considered. With a total of 10 energy landscapes, we can smoothly transition from a Gaussian-like landscape (with $K=10$) to a sharp and discontinuous landscape (with $K=1$).} $E^k_\theta(\vx, \vy)$ over the ground truth label distribution $p(\vy^* | x)$, where each energy function is learned to represent an EBM distribution
\begin{equation}
e^{-E^k_\theta(\vx, \vy)} \propto \int_{\vy^*} p(\vy^* | x) \cdot \mathcal{N}(\vy; \sqrt{1-\sigma_k^2} \vy^*, \sigma^2_k \mathbf{I}) 
\end{equation}
over a sequence of noise scales $\sigma_k$. Here, $\mathcal{N}(\cdot | \mu, \sigma)$ is the Gaussian density function. Larger values of $\sigma_k$ correspond to smoother energy landscapes while smaller values lead to sharper landscapes, with the energy minima of landscape $k$ corresponding to $\sqrt{1-\sigma_k^2} \vy^*$ (which can be scaled by $\tfrac{1}{\sqrt{1-\sigma_k^2}}$ to obtain the ground truth prediction $\vy^*$).  

We can directly learn each energy landscape by supervising the gradient of energy function to denoise the corrupted ground truth label $\vy^*$ from the dataset
\begin{equation}
    \label{eqn:denoise_diffusion}
      \mathcal{L}_{\text{MSE}}(\theta) = \| \nabla_{\vy} E_\theta(\vx, \sqrt{1-\sigma_k^2} \vy^* + \sigma_k \epsilon; k) - \epsilon \|^2, 
\end{equation}
where $\epsilon \sim \mathcal{N}(0, 1)$. Given a set of $K$ different learned energy landscapes, we can initialize a data sample from Gaussian noise and sequentially run $T$ steps of optimization following \eqn{eqn:energy_iterative} over each energy landscape $k$ (starting with high noise levels and progressing to lower noise levels). The optimization result in the previous energy landscape is used to initialize optimization in the next landscape, after scaled by the appropriate scaling factor $\tfrac{\sqrt{1-\sigma_k^2}}{\sqrt{1-\sigma_{k-1}^2}}$.

\subsection{Shaping the Energy Landscape}
\label{sect:energy_shape}
\vspace{-5pt}

In the denoising training objective \eqn{eqn:denoise_diffusion}, while the gradient of the energy landscape is locally trained to restore the ground truth label $\vy$, it is not necessarily the case that the overall global energy minima $\argmin_{\vy} E_\theta(\vx, \vy, k)$ corresponds to the ground truth label $\sqrt{1-\sigma_k^2} \vy^*$. 

To enforce that the global energy minima of each of the $k$ energy landscapes corresponds to the ground truth energy minima, we further propose a contrastive loss, where we construct a set of negative label $\vy^{-}$ (formed by noise corrupting the ground truth label $\vy^*$). Given an energy $E^{+} = E_\theta(\vx, \sqrt{1-\sigma_k^2} \vy^* + \sigma_k \epsilon; k)$ of the ground truth label $\vy^*$ and an energy $E_{-} = E_\theta(\vx, \sqrt{1-\sigma_k^2} \vy^{-} + \sigma_k \epsilon; k)$ of the negative label $\vy^{-}$, $\mathcal{L}_{\text{Contrast}}(\theta) = -\log \left ( \frac{e^{-E^+}}{e^{-E^+} + e^{-E^-}} \right)$.
To reduce the variance of the contrastive loss, we use the same sampled noise value $\epsilon$ for both $\vy$ and $\vy^{-}$.

\subsection{Combined Training and Inference Paradigms}
\label{sect:psuedocode}
\vspace{-5pt}

We provide the overall pseudocode for training \model in Algorithm \ref{alg:train} and executing algorithmic reasoning with \model in Algorithm \ref{alg:test}. We use a cosine beta schedule to train annealed energy landscapes and use a total of 10 energy landscapes (we empirically found that more landscapes did not lead to improved performance). At inference time, we can vary the number of optimization steps $T$ for each energy landscape to make trade-offs between performances and inference speed.

In principle, when the solution is not well-defined, it is possible to use IRED to model multi-modal distributions, similar to how diffusion models have been proven effective in modeling multi-modal image distributions. Depending on the particular use case, one may also add additional inference-time constraints (\eg, by composing the learned IRED energy function with other energy functions) to select favorable solutions.

\section{Experiments}

We compare \model with both domain-specific and domain-independent baselines on three domains: continuous algorithmic reasoning, discrete-space reasoning, and planning.
As we will break down in the following sections, the main advantages of IRED are twofold. First, compared with energy-based models (IREM), it is faster to train since it does not require backpropagation through the optimization process. Second, in terms of task performance, our focus is on generalization to ``harder'' problems, particularly leveraging the contrastive energy supervision and runtime iterative refinements. The idea is that after learning a correct energy landscape, the model can adaptively use more computation at test time to directly generalize to harder problems: we will focus on evaluating this generalization across all domains.

\subsection{Continuous Algorithmic Reasoning}
\label{sect:continuous}
\begin{table}[t]
\small\setlength{\tabcolsep}{5.5pt}
\centering
\begin{tabular}{llcc}
\toprule
      & & {\bf Same} & {\bf Harder} \\
      {\bf Task} & {\bf Method} & {\bf Difficulty} & {\bf Difficulty} \\
      \midrule
     \multirow{5}{*}{{\bf Addition}} & Feedforward & 0.0448  & 0.7029  \\
       & Recurrent & 0.3610 & 2.6133  \\
       & Programmatic &  0.0111 & 0.3446   \\
      & Diffusion & 0.0071 & 0.5931  \\
      & IREM & 0.0003 & 0.0021  \\
      & \model (ours) & \textbf{0.0002} & \textbf{0.0020}  \\
      
      \midrule
     \multirow{4}{*}{ {\bf Matrix}} & Feedforward & 0.0203 & 0.2720   \\
      \multirow{4}{*}{ {\bf Completion}} & Recurrent & 0.0266 & 0.3285 \\
      & Programmatic &  0.0203 & 0.2637  \\
      & Diffusion  & 0.0219 & 0.2142 \\
      & IREM & 0.0183 & 0.2074 \\
      & \model (Ours)  & \textbf{0.0174} & \textbf{0.2054}\\
      \midrule
      \multirow{4}{*}{ {\bf Matrix}} & Feedforward & 0.0112 & 0.2150   \\
      \multirow{4}{*}{ {\bf Inverse}} & Recurrent & 0.0109 &   0.2123  \\
      & Programmatic &   0.0124  & 0.2209 \\
      & Diffusion & 0.0115 &  0.2132 \\
      & IREM & 0.0108 & 0.2083 \\
      & \model (Ours)  & \textbf{0.0095} & \textbf{0.2063}\\
    \bottomrule
\end{tabular}
\caption{\textbf{Continuous  Algorithmic Reasoning.} Test evaluation performance on continuous algorithmic tasks. Inputs and outputs are 20 by 20 matrices. Error is reported using elementwise mean square error. Models are evaluated on test problems drawn from the training distribution (same difficulty) and a harder test distribution (harder difficulty). \model outperforms comparisons.}
\label{tbl:tbl_continuous}
\vspace{-5pt}
\end{table}

\myparagraph{Setup.} We first evaluate \model on a set of continuous algorithmic reasoning tasks from ~\citet{du2022learning}. We consider three matrix operations on $20 \times 20$ matrices, which are encoded 400-dimensional vectors:
\vspace{-5pt}
\begin{enumerate}
    \myitem \textit{Addition}: We first evaluate neural networks in their ability to add matrices together (element-wise). We also evaluate neural network on harder variants of the addition problems at test time by feeding input vectors with larger magnitudes. 
    \myitem \textit{Matrix Completion}: Next, we evaluate neural networks on their ability to do low-rank  matrix completion. We mask out $50\%$ of the entries of a low-rank input matrix constructed two separate rank 10 matrices $U$ and $V$, and train networks to reconstruct the original input matrix. We construct harder variants of the matrix completion problem at test time by increasing the magnitude of values in $U$ and $V$. 
    \myitem \textit{Matrix Inverse}: Finally, we evaluate neural networks on their ability to compute matrix inverses. We construct harder matrix inverse problems by considering less well-conditioned input matrices. 
\end{enumerate}
\vspace{-10pt}

We report the underlying mean-squared error (MSE) between the predictions and the associated ground truth outputs on test problem instances. To more effectively generate negative samples for \model in this domain, we first noise-corrupt ground truth labels and then run two steps of gradient optimization on the energy landscape to form negative samples. Details can be found in Appendix~\ref{sect:experimental_detail}.  

\myparagraph{Baselines.} We compare our approach to a set of iterative reasoning baselines found in ~\citep{du2022learning}: \textit{(Feedforward):} an iterative reasoning approach where the same MLP is repeatedly applied, \textit{(Recurrent):} an iterative reasoning approach where the recurrent network is repeatedly applied, and \textit{(Programmatic):} an iterative reasoning approach which repeatedly applies a learned programmatic module~\citep{banino2021pondernet}.  We further compare with the IREM method ~\citep{du2022learning}, as well as using a denoising diffusion model directly to solve continuous tasks. All methods use identical architectures (with small differences due to recurrent layers or timestep conditioning).

\myparagraph{Quantitative Results.} We compare \model with baselines across settings in \tbl{tbl:tbl_continuous}. Similar to IREM, \model is able to nearly perfectly solve the task of the addition, as well as generalize to larger addition matrices. On other tasks, \model outperforms IREM and also generalizes better to harder problems. Furthermore, our approach substantially outperforms directly using a diffusion process to predict solutions, which lacks the iterative energy minimization procedure that explicitly learns the task constraints.

\begin{figure}[t]
\vspace{-5pt}
\includegraphics[width=\linewidth]{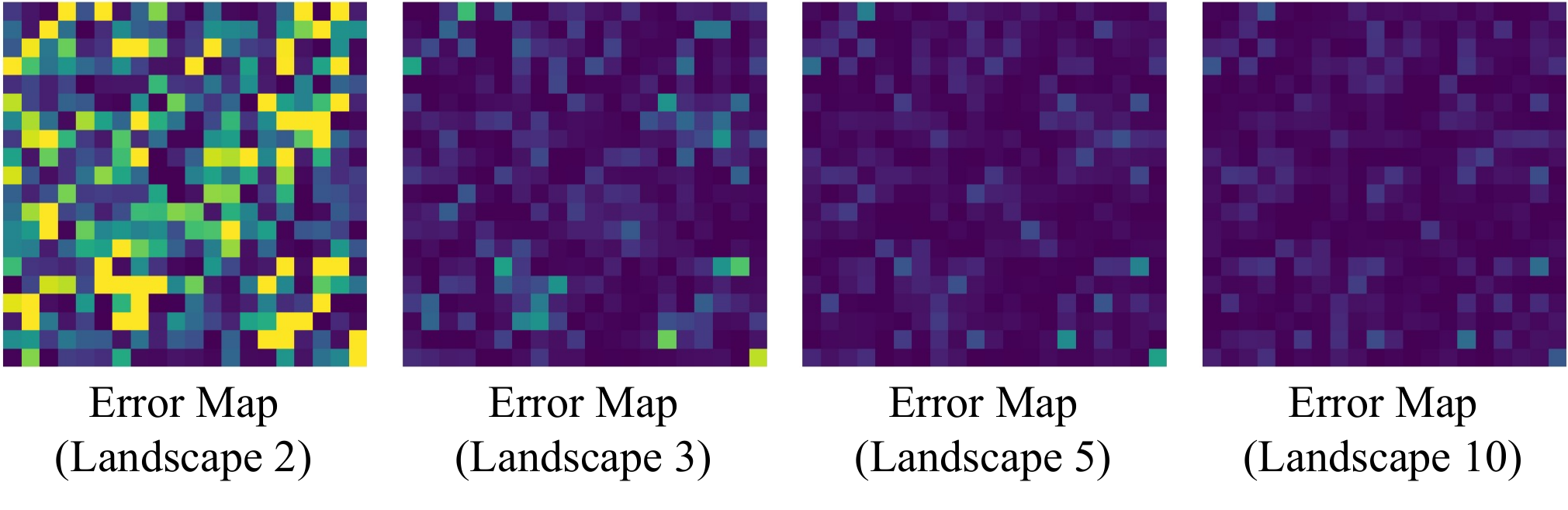}
\vspace{-25pt}
\caption{\small \textbf{Optimized Solutions Across Landscapes} -- Error maps of intermediate optimized solutions. Optimized solutions at earlier landscapes are less accurate than later ones.}
\label{fig:energy_minima}
\vspace{-5pt}
\end{figure}

\begin{figure}[t]
\vspace{-5pt}
\includegraphics[width=\linewidth]{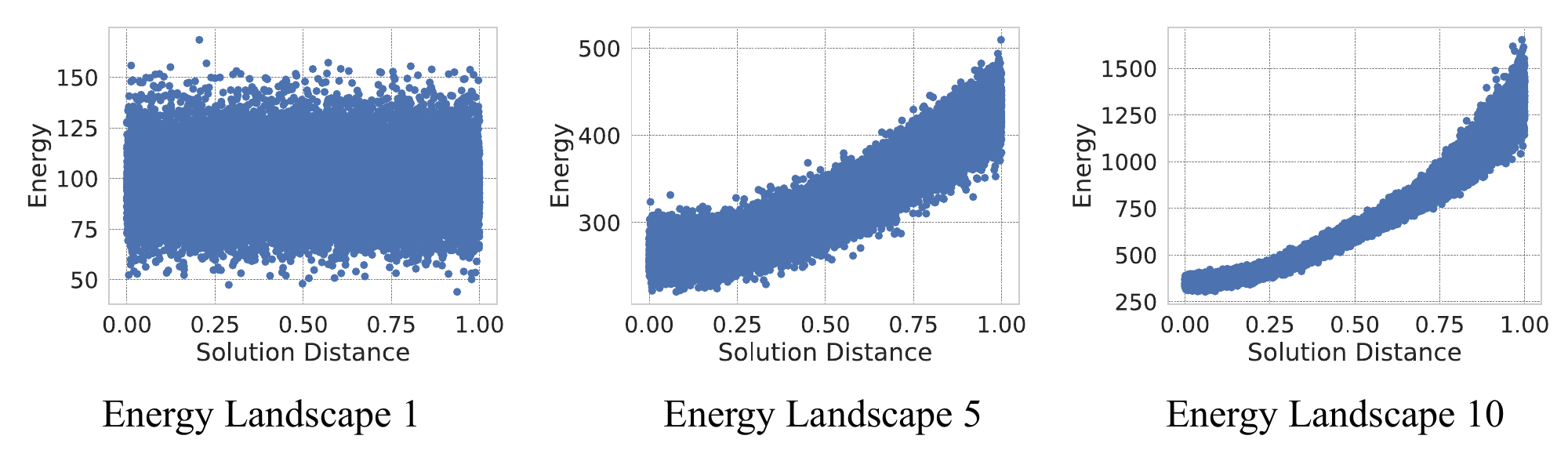}
\vspace{-20pt}
\caption{\small \textbf{Energy Landscape} -- Predicted energy values for $\vy$ and the corresponding MSE distance of $\vy$ from the problem solution across different landscapes on the matrix inverse task. The earlier energy landscapes are smoother than the later ones.}
\label{fig:energy_landscape}
\vspace{-10pt}
\end{figure}

\begin{table}[tp]
\small\setlength{\tabcolsep}{5.5pt}
\centering\small
\begin{tabular}{lcc}
\toprule
      {\bf Opt. Steps} & {\bf Same Difficulty} & {\bf Harder Difficulty} \\
      \midrule
      10 & 0.0096 & 0.2110 \\
      20 & 0.0096 & 0.2100 \\
      30 & 0.0096 & 0.2090\\
      40 & 0.0095 & 0.2063 \\
    \bottomrule
\end{tabular}
\vspace{-5pt}
\caption{\small \textbf{Continuous-Space Reasoning Performance vs Reasoning Steps.} More reasoning steps in \model at inference time substantially improve generalization to harder difficulty tasks on the matrix inverse task. \model is trained with 10 energy landscapes. }
\label{tbl:ablation_step_size}
\vspace{-5pt}
\end{table}

\begin{table}[t]
    \centering

\setlength{\tabcolsep}{3pt}
    \scalebox{0.85}{
    \begin{tabular}{ccccc}
        \toprule
        \textbf{Gradient}  & \textbf{Optimization} & \textbf{Contrastive} & \textbf{Same} & \textbf{Harder} \\
         \textbf{Descent} & \textbf{Refinement} & \textbf{Shaping} & \textbf{Difficulty} & \textbf{Difficulty} \\
        \midrule
        No & No & No & 0.0158 & 0.2223 \\
        Yes & No & No & 0.0097 & 0.2135 \\
        Yes & Yes & No  & 0.0097 & 0.2113 \\
        Yes & Yes & Yes  & 0.0095 & 0.2063 \\
        \bottomrule
    \end{tabular}
    }
    \vspace{-5pt}
    \caption{\small \textbf{Continuous Ablations --} Ablations of proposed components of \model on performance on the matrix inverse task. Leveraging gradient descent to optimize energy functions, using multiple steps of optimization at each energy landscape and contrastively shaping the energy landscape with ground truth labels all improve the performance on the Inverse task. }
     \label{tbl:ablation}
    \vspace{-5pt}
\end{table}

\myparagraph{Qualitative Visualization.} We provide a qualitative visualization of the error map of the optimized solution on the matrix inverse task at each different learned energy landscape in \fig{fig:energy_minima}. The error of optimized solutions at different energy landscapes decreases over time.

\myparagraph{Energy Landscape.} We visualize the learned energy landscape in \fig{fig:energy_landscape} as a function of the distance of an input label from the ground truth label. In early energy landscapes, the difference between energy values of solutions close and far from the ground truth solution is low, and therefore the energy landscape is relatively flat. At later landscapes, the energy value increases substantially as the input solution deviates from the ground truth solution.

\myparagraph{Performance with Increased Computation.} We analyze the performance on the matrix inverse task as a factor of an increased number of computational steps in \tbl{tbl:ablation_step_size}. We find that running additional steps of optimization slightly improves performance on in-distribution tasks and substantially improves performance on harder problems.

\myparagraph{Ablation.} We ablate each component of \model in \tbl{tbl:ablation}. In the first two rows of \tbl{tbl:ablation}, we compare our gradient-descent-based optimization with a noisy optimization procedure corresponding to the diffusion reverse process for each energy landscape. In the third row, we compare the difference between running multiple steps of optimization as opposed to a single energy optimization step. Finally, We then consider the effect of contrastively shaping the energy landscape. All components lead to improved performance.

\vspace{-3pt}
\subsection{Discrete-Space Reasoning}

\myparagraph{Setup.} The second group of tasks evaluates \model on its reasoning in discrete spaces (\ie, values are all binary or one-hot categorical). We run evaluations on two tasks: Sudoku solving and graph connectivity reasoning.

\vspace{-5pt}
\begin{enumerate}
    \myitem \textit{Sudoku}: In the Sudoku game, the model is given a partially filled Sudoku board, with 0's filled-in entries that are currently unknown. The task is to predict a valid solution that jointly satisfies the Sodoku rules and that is consistent with the given numbers. We use the dataset from SAT-Net~\citep{wang2019satnet} as the training and standard test dataset. In SAT-Net, the number of given numbers is within the range of $[31, 42]$. Our harder dataset is from RRN~\citep{palm2018recurrent}, which is a different Sudoku dataset where the number of given numbers is within $[17, 34]$. We will show that our system, being trained on simpler Sudoku games with fewer blank entries, generalizes to harder instances.
    \myitem \textit{Connectivity}: In the graph connectivity task, the model is given the adjacency matrix of a graph (1 if there is an edge directly connecting two nodes). The task is to predict the connectivity matrix of the graph (1 if there is a path connecting two nodes). In literature~\citep{dong2019neural}, this task is an example that requires a dynamic number of reasoning ``steps'' (depending on the treewidth of the graph). Therefore, prior papers primarily focus on computing connectivity between nodes within $k$-steps away. Our training and standard test sets contain graphs with at most 12 nodes and our harder dataset contains graphs with 18 nodes. Since we do not limit the path lengths between nodes (in contrast to \citet{dong2019neural}), on the training set, the maximum distance between two connected nodes is 9, and in the harder test set, the maximum distance is 16.
\end{enumerate}
\vspace{-7pt}

\begin{table}
\small\setlength{\tabcolsep}{5.5pt}
\centering
\begin{tabular}{llcc}
\toprule
      {\bf Task} & {\bf Method} & {\bf Test} & {\bf Harder} \\
      & & {\bf Dataset} & {\bf Dataset} \\
      \midrule
      \multirow{4}{*}{\bf Sudoku} & IREM & 93.5\% & 24.6\%  \\
      & Diffusion & 66.1\% & 10.3\% \\
      & SAT-Net & 98.3\% & 3.2\% \\
      & RRN & {\bf 99.8\%} & 28.6\% \\
      & \model (ours) & 99.4\% & {\bf 62.1\%}  \\
      \midrule
      \multirow{2}{*}{\bf Visual Sudoku} 
      & SAT-Net & 63.2\% & 0.0\% \\
      & RRN & {\bf 99.8\%} & 28.6\% \\
      & \model (ours) &  98.3\% & {\bf 46.6\%}  \\  
      \midrule
      \multirow{3}{*}{\bf Connectivity} & IREM & 94.3\% &  89.8\% \\
      & Diffusion & 61.6\% & 61.3\% \\
      & \model (ours) & {\bf 99.1\%} & {\bf 93.8\%}  \\  
    \bottomrule
\end{tabular}
\caption{\textbf{Discrete Reasoning Performance.} We evaluate models on the Sudoku task and the connectivity task. Sudoku:" the harder dataset has between 17 to 34 entries given, while models are trained with 31 to 42 entries given. Connectivity: the harder graphs have at most 18 nodes while the training graphs have only 12.}
\label{tbl:tbl_discrete}
\vspace{-5pt}
\end{table}

\begin{figure}[t]
\includegraphics[width=\linewidth]{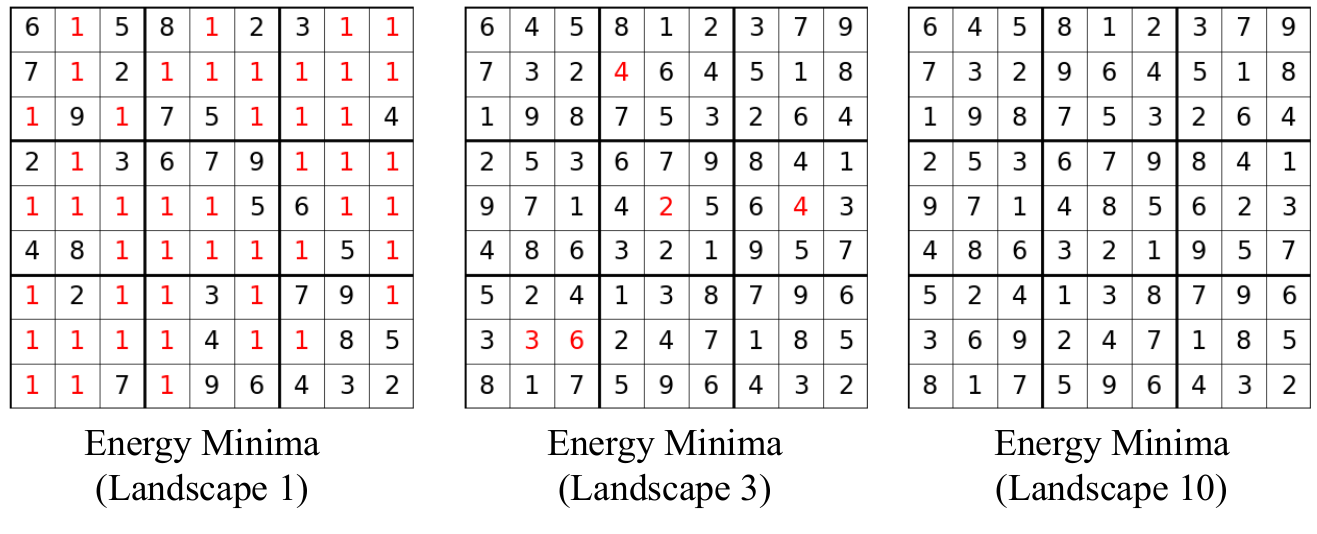}
\vspace{-25pt}
\caption{\small \textbf{Optimized Boards Across Landscapes} -- Plot of the minimal energy board across energy landscapes, given the same initial board. Later energy landscapes lead to more accurate boards. We highlight inconsistent entries in red.}
\label{fig:sudoku_opt}
\vspace{-15pt}
\end{figure}

\begin{figure}[t]
\vspace{-5pt}
\centering
\includegraphics[width=0.8\linewidth]{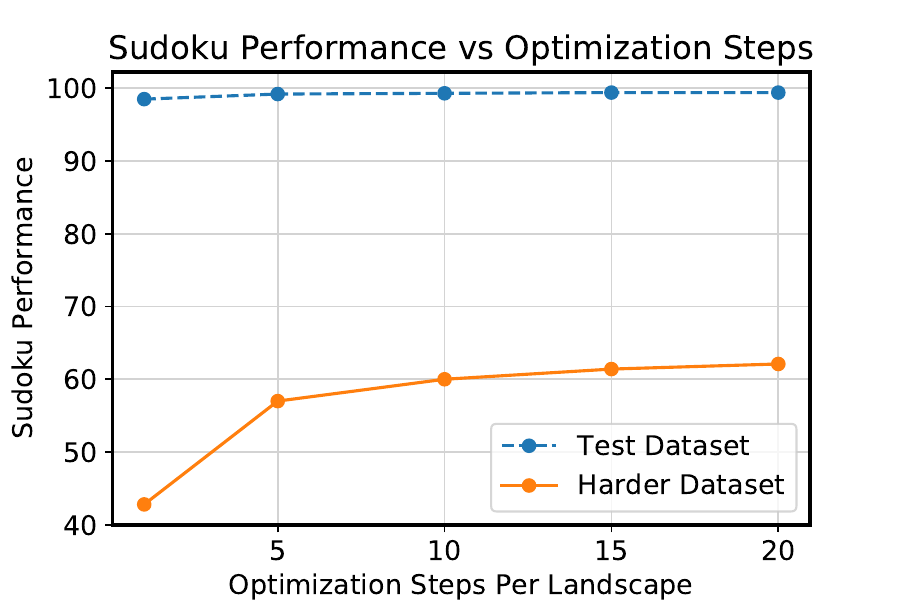}
\vspace{-5pt}
\caption{\small \textbf{Sudoku Performance with Optimization Steps} -- Generalization to harder Sudoku problems significantly improves with a larger number of optimization steps in Sudoku.}
\label{fig:sudoku_performance_step}
\end{figure}

\begin{table}[t]
    \centering

\setlength{\tabcolsep}{3pt}
    \scalebox{0.85}{
    \begin{tabular}{ccccc}
        \toprule
        \textbf{Gradient}  & \textbf{Optimization} & \textbf{Contrastive} & \textbf{Same} & \textbf{Harder} \\
         \textbf{Descent} & \textbf{Refinement} & \textbf{Shaping} & \textbf{Difficulty} & \textbf{Difficulty} \\
        \midrule
        No & No & No & 97.0\% & 35.0\% \\
        Yes & No & No & 98.8\% & 45.1\% \\
        Yes & Yes & No  & 99.3\%  & 59.7\% \\
        Yes & Yes & Yes  & 99.4\%  &  62.1\% \\
        \bottomrule
    \end{tabular}
    }
  
    \caption{\small \textbf{Discrete Reasoning Ablations --} Ablations of proposed components of \model on performance on the Sudoku task. Leveraging gradient descent to optimize energy functions, using multiple steps of optimization at each energy landscape and contrastively shaping the energy landscape with ground truth labels all improve the performance. }
     \label{tbl:discrete_ablation}
    \vspace{-10pt}
\end{table}

\myparagraph{Quantitative Results.} We compare \model with both IREM and diffusion baselines. On the Sudoku task, we further compare with the domain-specific SAT-Net method. In \tbl{tbl:tbl_discrete}, we find that our approach substantially outperforms all baselines across both evaluated discrete-space reasoning settings. In Sudoku, our approach generalizes substantially better than the SAT-Net model and the RRN model to the harder dataset consisting of fewer given Sudoku elements and is capable of obtaining an accuracy of roughly 62.1\% compared to an accuracy of 3.2\% obtained by SAT-Net and 28.6\% by RRN. For cases in which our approach fails, we found that our approach sometimes erroneously assigns low energy to partially accurate answers. For example, there can be a Sudoku board that is not fully valid, but the model assigns lower energy to it than to the ground truth board.

\myparagraph{Qualitative Results.} We qualitatively visualize intermediate optimized samples across energy landscapes on Sudoku in \fig{fig:sudoku_performance_step}. Optimized boards are increasingly accurate in later landscapes.

\myparagraph{Performance with Computation.} In \fig{fig:sudoku_performance_step}, we assess the impact of the number of optimization steps at each energy landscape on the performance in Sudoku on both the test and harder datasets. We find that performance substantially improves on the harder dataset with an increased number of optimization steps with more modest improvement on the test datasets. By formulating reasoning as energy optimization, we can adaptively change the number of optimization steps dependent on difficulty of task, enabling us to generalize substantially better on harder tasks.

\myparagraph{Extension to Visual Sudoku.} \model can also be extended to deal with other input formats, such as images. To illustrate this, we conducte a new experiment on the Visual Sudoku dataset~\citep{wang2019satnet}, where the board is not represented by one-hot vectors but now consists of MNIST digits written on a grid. We use a CNN to encode the image and fuse the image embeddings with the noisy answer to predict energy values. Shown in Table~\ref{tbl:tbl_discrete}, we observed a similar performance advantage of our model compared to the baseline.

\myparagraph{Ablation.} In \tbl{tbl:discrete_ablation}, we ablate each component of \model on the Sudoku task. Similar to the continuous setting, we find that each component of \model, gradient based optimization, multiple steps of optimization, and contrastive energy shaping all lead to improved performance. While performance is modestly improved on the test dataset, it is substantially improved on the harder generalization dataset with each added component.

\subsection{Planning}

\myparagraph{Setup.} In this section, we evaluate \model on a basic decision-making problem of finding the shortest path in a graph. In this task, the input to the model is the adjacency matrix of a directed graph, together with two additional node embeddings indicating the start and the goal node of the path-finding problem. The task is to predict a sequence of actions corresponding to the plan. Concretely, the output is a matrix of size $[T, N]$, where $T$ is the number of planning steps and $N$ is the number of nodes in the graph. Each entry $(t, i)$ has a value of 1 if the $t$-th step of the shortest path is at node $i$ and has a value of 0 otherwise. For all models, including ours, we use a spatial-temporal graph convolution network~\citep[STGCN;][]{yan2018spatial} to encode the adjacency matrix and the prediction and predict energy function values or score functions. In short, the STGCN has multiple layers. At each layer, for each node, it fuses all features of nodes that are connected to it in the graph and from the current timestep or adjacent timesteps. Just like standard graph convolutional networks, it uses a sum-pooling mechanism to aggregate all embeddings from adjacent nodes. Since, in practice, such planning models are generally evaluated with closed-loop execution, here we only evaluate the success rate that the first action produced by the model shortens the distance between the current node and the goal node. This is the same planning and execution strategy as DiffusionPolicy~\citep{chi2023diffusion}.

\begin{table}
\small\setlength{\tabcolsep}{5.5pt}
\centering
\begin{tabular}{llcc}
\toprule
      {\bf Task} & {\bf Method} & {\bf Test} & {\bf Harder} \\
      & & {\bf Dataset} & {\bf Dataset} \\
      \midrule
      \multirow{3}{*}{\bf Shortest Path} & IREM & 90.4\% & 88.4\% \\
      & Diffusion & 45.2\% & 46.9\% \\
      & \model (ours) & {\bf 92.6\%} & {\bf 91.9\%}  \\  
    \bottomrule
\end{tabular}
\vspace{-5pt}
\caption{\textbf{Planning Performance.} Test evaluation performance on the shortest path task. The harder tasks consists of graphs of size 25 while models are trained on graphs of size 15.}
\label{tbl:tbl_planning}
\vspace{-5pt}
\end{table}

\begin{figure}[t]
\includegraphics[width=\linewidth]{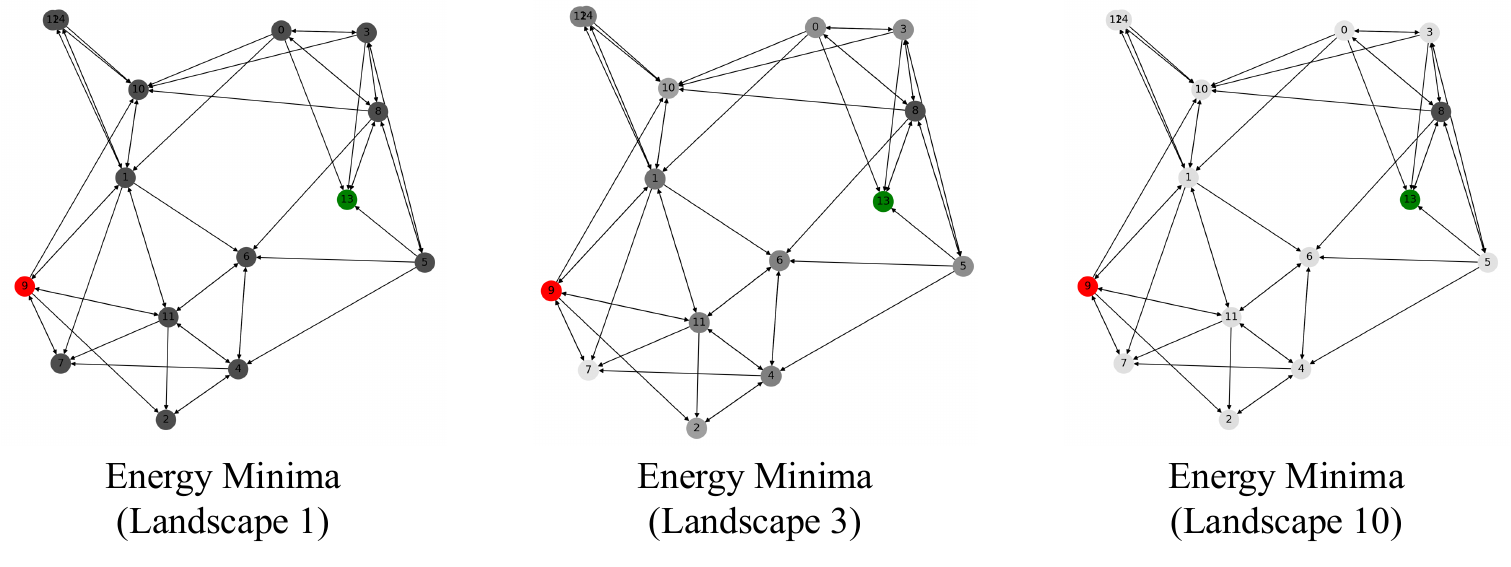}
\vspace{-15pt}
\caption{\small \textbf{Optimized Plans Across Landscapes} -- Plot of next action prediction in plans across energy landscapes. In each visualization, the green/red nodes indicate start/goal nodes with connections between nodes indicated with arrows. The darkness of a node indicates the score for selecting the corresponding node as the next node to move to in the predicted plan. As landscapes are sequentially optimized, the correct next action is selected.}
\label{fig:path_planning_opt}
\vspace{-15pt}
\end{figure}

\begin{table}[t]
    \centering

\setlength{\tabcolsep}{3pt}
    \scalebox{0.85}{
    \begin{tabular}{ccccc}
        \toprule
        \textbf{Gradient}  & \textbf{Optimization} & \textbf{Contrastive} & \textbf{Same} & \textbf{Harder} \\
         \textbf{Descent} & \textbf{Refinement} & \textbf{Shaping} & \textbf{Difficulty} & \textbf{Difficulty} \\
        \midrule
        No & No & No & 80.8\% & 80.4\% \\
        Yes & No & No & 80.7\% & 79.1\% \\
        Yes & Yes & No  & 88.6\%  & 87.9\% \\
        Yes & Yes & Yes  & {\bf 92.6\%}  &  {\bf 91.9\%} \\
        \bottomrule
    \end{tabular}
    }
\vspace{-5pt}
    \caption{\small \textbf{Path Planning Ablations --} Ablations of proposed components of \model on performance on the path planning task. Using multiple steps of optimization at each energy landscape and contrastively shaping the energy landscape with ground truth labels both improve path planning performance. }
     \label{tbl:path_ablation}
    \vspace{-10pt}
\end{table}

\myparagraph{Quantitative Results.} We compare \model with all baselines across settings in \tbl{tbl:tbl_planning}. \model outperforms both baselines, especially the diffusion model, by a large margin. Both methods based on energy based formulation (IREM and \model) perform well on this task, validating the hypothesis that learning energy functions is an effective method for encoding planning problems such as the edge constraints in the path-finding task. Finally, since all methods uses the same STGCN encoder, there is no significant performance drop on generalization to the harder dataset.

\myparagraph{Qualitative Results.} Similarly to other tasks, we can also visualize the generated solutions by the model across different landscapes. \fig{fig:path_planning_opt} visualizes the prediction of the first node to move to by the planning model. We normalize the prediction scores to 0 to 1. The darker the color, the higher the score. As can be seen in the figure, our \model model is capable of gradually finding the immediate next action to take: at step 3 it excluded node 7, and the score gradually concentrates on node 8.

\myparagraph{Ablations.} We ablate each component of \model in \tbl{tbl:path_ablation}. We found that running multiple steps of optimization and shaping the energy landscape both lead to improved performance on both the same-difficulty test cases and harder difficulty ones.

\section{Conclusion and Discussions}

In this paper, we present \model, a new approach to solving complex reasoning tasks by formulating it as an energy minimization process on a sequence of energy landscapes. We illustrate, on both continuous, discrete, and planning domains, how iterative computation utilizing \model enables better algorithmic performance and generalization to more complex instances of problems. We further illustrate how the underlying algorithmic computation learned by \model may be nested to implement more complex algorithmic computations. 

Our current reasoning approach with \model has several limitations. First, the inference time optimization procedure in \model can still be improved because currently, it requires many steps of gradient descent to find an energy minima. For tasks with known specifications (e.g., shortest path), IRED will conceivably run slower than the algorithms designed specifically for them (e.g., polynomial algorithms for pathfinding), although it is worth noting that IRED is a general machine learning algorithm that does not assume a given task specification and can automatically learn the underlying task constraints from data. On the other hand, it would be interesting to explore how an amortized neural network generator for generating initial solutions or guided optimizers can speed up this procedure. Second, our sequence of annealed energy landscapes is defined through a sequence of added Gaussian noise increments --- it would be further interesting to learn the sequence of energy landscapes to enable adaptive optimization. Another current limitation of \model is that out of the box, IRED in its current form does not leverage any additional memory. Therefore, for tasks that would benefit from explicitly using additional memory to store intermediate results (analogous to chain-of-thought reasoning tasks in language or visual reasoning), IRED might not be as effective as other approaches.

So far we have been applying \model in continuous and discrete reasoning tasks, and planning on discrete spaces (graphs). Other potential applications of \model include general mathematical reasoning~\citep{amini-etal-2019-mathqa,lu2021inter} and decision-making in hybrid discrete-continuous spaces~\citep{caelan2021tamp,yang2023diffusion,fang2023dimsam}. For example, IRED can serve directly as a policy model. In our experiments, we have supervised the energy functions using IID samples from a fixed dataset (as positive examples). However, the energy function can also be supervised with reward signals by modeling a distribution of actions that favor actions yielding high discounted returns (similar to REINFORCE). 

\vspace{-2pt}
\section*{Acknowledgements}
We gratefully acknowledge support from NSF grant 2214177; from AFOSR grant FA9550-22-1-0249; from ONR MURI grant N00014-22-1-2740; and from ARO grant W911NF-23-1-0034; from MIT Quest for Intelligence; from the MIT-IBM Watson AI Lab; from ONR Science of AI; and from Simons Center for the Social Brain. Yilun Du is supported by a NSF Graduate Fellowship. Any opinions, findings, and conclusions or recommendations expressed in this material are those of the authors and do not necessarily reflect the views of our sponsors.

\vspace{-2pt}
\section*{Impact Statement}
No immediate negative social impacts can be derived from the proposed approach in its current form as our evaluation is carried out on relatively standard simple datasets.

\bibliography{references}
\bibliographystyle{icml2024}

\clearpage
\twocolumn[
\vspace{2em}
\centering
\textbf{\Large Supplementary Material for Iterative Reasoning through Energy Diffusion}
\vspace{2em}
]
\appendix

In this appendix we provide additional details on \model. We first provide experimental details on our evaluated tasks in Appendix~\ref{sect:experimental_detail}. Next, we discuss individual model architectures used in Appendix~\ref{sect:model_architecture}.

\section{Experimental Details}
\label{sect:experimental_detail}

\myparagraph{Continuous Tasks} We use dataset setups from ~\citep{du2022learning} for continuous tasks. Models were trained in approximately 2 hours on a single Nvidia RTX 2080 using a training batch size of 2048 and the Adam optimizer with learning rate 1e-4. Models was trained for approximately 50,000 iterations and evaluated on 20000 test problems. 

\myparagraph{Discrete Tasks} For Sudoku, we train models for 50000 iterations using a single Nvidia RTX 2080 using a training batch size of 64 with the Adam optimizer with learning rate 1e-4 and are evaluated on the full test datasets provided in ~\citep{wang2019satnet, palm2018recurrent}. 

For Connectivity tasks, we generate random graphs using algorithms from \citet{graves2016hybrid}. Essentially, it first generates a set of random points on a 2D plane uniformly inside a unit square. Next, it samples the out-degree $k$ (the number of out-going edges) for each node based on a uniform distribution. Finally, it connects each node to its $k$ closest neighbors. The advantage of this generation process is that the generated graph will be close to a planar graph so that it can be easily visualized, and more importantly, it allows for fine-grained control of the connectivity of the graph by setting the out-degree range. In practice, we uniformly sample the out-degrees from $[1, n/2]$, where $n$ is the number of nodes in the graph. Based on the sampled adjacency matrix, we use the floyd-warshall algorithm to compute its connectivity matrix. Note that there are no distances associated with the edges (\ie, all edges are of unit length). We train models for 100000 iterations using a single Nvidia RTX 2080 with batch size 512 with the Adam optimizer.

\myparagraph{Planning Task} For planning, we use the same procedure as in the connectivity tasks to generate graphs. To create graphs with a large enough treewidth, we set the out-degree sample range to be $[1, n/5]$, where $n$ is the number of nodes. All edges are of unit length. We use the floyd-warshall algorithm to compute the pairwise shortest distance. This enables us to evaluate whether the first action predicted by the model shortens the distance between the current node and the goal. We train models for 100000 iterations using a single Nvidia RTX 2080 with batch size 512 with the Adam optimizer.

\section{Model Architectures}
\label{sect:model_architecture}

\myparagraph{Continuous Task.}  For continuous tasks, we use the architecture in \tbl{tbl:continuous_ebm} to train both IRED and the IREM baseline. We use the architecture in \tbl{tbl:continuous_ff} to parameterize the diffusion model baseline. 

\myparagraph{Discrete Task.} For Sudoku, we use the architecture in \tbl{tbl:sudoku_ebm}. It encodes the Sudoku board with a convolutional neural network with the residual connection design, borrowed from \citet{he2016deep}.

For Connectivity, we use the architecture adapted from \citet{dong2019neural}, as detailed in \tbl{tbl:connectivity_ebm}. It uses a relational neural network to fuse the connectivity information from neighboring nodes.

\myparagraph{Planning Task.}  For planning tasks, we use the architecture the same architecture as the connectivity task, as detailed in \tbl{tbl:shortest_path_ebm}. To encode temporal information, at each layer, we stack the node embedding from the previous time step, the current step, and the next step to implement temporal convolution. This is equivalent to the spatial-temporal graph convolution networks~\citep[STGCN;][]{yan2018spatial}.

We have also attached the code to reproduce all experiments in the supplementary material.

\clearpage

\begin{figure}[H]
\centering
\small
\begin{tabular}{c}
    \toprule
    Linear 512 \\
    \midrule
    Linear 512 \\
    \midrule
    Linear 512 \\
    \midrule
    Linear $\rightarrow$ 1 \\ 
    \bottomrule
\end{tabular}
\captionof{table}{The model architecture for \model on continuous-space algorithmic reasoning tasks}
\label{tbl:continuous_ebm}
\end{figure}

\begin{figure}[H]
\centering
\small
\begin{tabular}{c}
    \toprule
    Linear 512 \\
    \midrule
    Linear 512 \\
    \midrule
    Linear 512 \\
    \midrule
    Linear $\rightarrow$ Output Dim \\ 
    \bottomrule
\end{tabular}
\captionof{table}{The model architecture for diffusion baselines on continuous-space algorithmic reasoning tasks.}
\label{tbl:continuous_ff}
\end{figure}

\begin{figure}[H]
\begin{minipage}{0.95\linewidth}
\centering
\small
\begin{tabular}{c}
    \toprule
    3x3 Conv2D, 384 \\
    \midrule
    Resblock 384 \\
    \midrule
    Resblock 384 \\
    \midrule
    Resblock 384 \\
    \midrule
    Resblock 384 \\
    \midrule
    Resblock 384 \\
    \midrule
    Resblock 384 \\
    \midrule
    3x3 Conv2D, 9 \\ 
    \bottomrule
\end{tabular}
\captionof{table}{The model architecture for \model and diffusion baselines on the Sudoku task. The energy value is computed using the L2 norm of the final predicted output similar to ~\citep{du2023reduce}, while the output is directly used as noise prediction for the diffusion baseline.}
\label{tbl:sudoku_ebm}
\end{minipage}%
\end{figure}

\begin{figure}[H]
\begin{minipage}{0.95\linewidth}
\centering
\small
\begin{tabular}{c}
    \toprule
    NLM Arity=3, Hidden=64 \\
    \midrule
    NLM Arity=3, Hidden=64 \\
    \midrule
    Max-Pooling over All Edge Features \\
    \midrule
    Linear, 1 \\ 
    \bottomrule
\end{tabular}
\captionof{table}{The model architecture for \model and diffusion baselines on the connectivity task. For the diffusion baseline, we simply remove the pooling layer and apply the same linear layer on all edge embeddings to predict the noise value for each entry.}
\label{tbl:connectivity_ebm}
\end{minipage}%
\end{figure}

\begin{figure}[H]
\begin{minipage}{0.95\linewidth}
\centering
\small
\begin{tabular}{c}
    \toprule
    NLM Arity=2, Hideen=64 \\
    \midrule
    NLM Arity=2, Hideen=64 \\
    \midrule
    Max-Pooling over All Node Features \\
    \midrule
    Linear, 1 \\ 
    \bottomrule
\end{tabular}
\captionof{table}{The model architecture for \model and diffusion baselines on the shortest-path task. For the diffusion baseline, we simply remove the pooling layer and apply the same linear layer on all node embedding to predict the noise value for each entry.}
\label{tbl:shortest_path_ebm}
\end{minipage}%
\end{figure}

\end{document}